\documentclass{article}

     \usepackage[final]{neurips_2020}

\usepackage[utf8]{inputenc} 
\usepackage[T1]{fontenc}    
\usepackage{hyperref}       
\usepackage{url}            
\usepackage{booktabs}       
\usepackage{amsfonts}       
\usepackage{nicefrac}       
\usepackage{microtype}      
\usepackage{graphicx}
\usepackage{subcaption}
\usepackage[title]{appendix}
\usepackage{color} 

\title{Image Generation With Neural Cellular Automatas}

\author{%
  Mingxiang Chen \\
  Beike \\
  Beijing, China \\
  \texttt{chenmingxiang002@ke.com} \\
   \And
  Zhecheng Wang \\
  Stanford University \\
  Stanford, CA 94305 \\
  \texttt{zhecheng@stanford.edu} \\
}

\begin{document}

\maketitle

\newcommand{\zhecheng}[1]{{\textcolor{black}{#1}}}

\begin{abstract}
In this paper, we propose a novel approach to generate images (or other artworks) by using neural cellular automatas (NCAs). Rather than training NCAs based on single images one by one, we combined the idea with variational autoencoders (VAEs), and hence explored some applications, such as image restoration and style \zhecheng{fusion.} 
The code for model implementation is available online \footnote{https://github.com/chenmingxiang110/VAE-NCA}.
\end{abstract}

\section{Introduction}

Cellular automata (CA) is a discrete model, which has been \zhecheng{widely} studied in mathematics, computational science, and biology. It is composed of an infinite number of regular squares, each of which is in a finite state. The entire grid can be any discrete finite dimension. The "neighbors" of each grid have been fixed. Every time the system evolves, each grid evolves \zhecheng{following} 
the same rules.

In neural cellular automata (NCA), the state of each grid is usually represented by a vector. The values in the vector can represent some hidden information, as well as color, \zhecheng{class, }
and other information. This method has been proven to be used for single image reconstruction [\cite{mordvintsev2020growing}]. However, the potential of this method should not stop there. It can completely replace the up-sampling part of many image generation models. Therefore, in this article, we combine NCA and VAE (VAE-NCA) to demonstrate their performance in image generation tasks such as style \zhecheng{fusion.}

\begin{figure}[h]
    \centering
    \includegraphics[width=.95\linewidth]{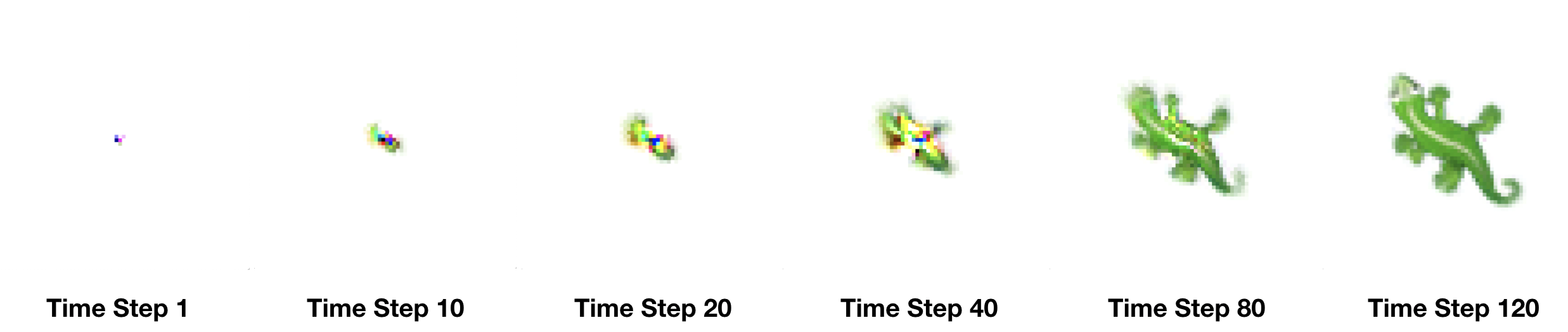}
    \caption{An example of the growing neural cellular automata [\cite{mordvintsev2020growing}] (our implementation) at different time steps.}\label{Fig:intro_NCA_example}
\end{figure}

This paper is organized as follows. First, we introduced the experimental results in Section 2. Then the ethical implications are discussed in Section 3. Due to the page limit, we describe our method and other training parameters in detail in the appendix.

\section{Experimental Results}

\begin{figure}[t!]
    \centering
    \begin{subfigure}[t!]{0.6\linewidth}
        \includegraphics[width=.99\linewidth]{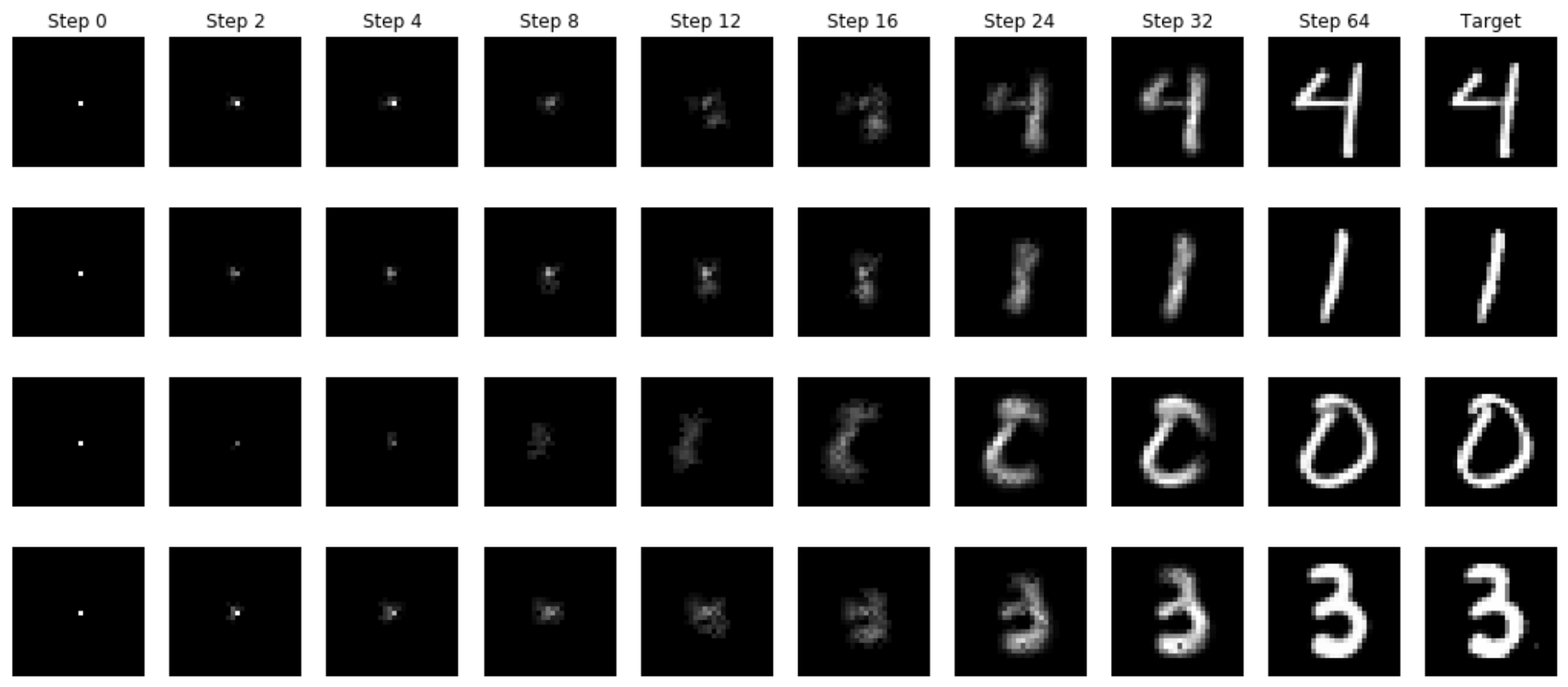}
        \caption{MNIST}
    \end{subfigure}
    \hfill
    \begin{subfigure}[t!]{0.38\linewidth}
        \includegraphics[width=.99\linewidth]{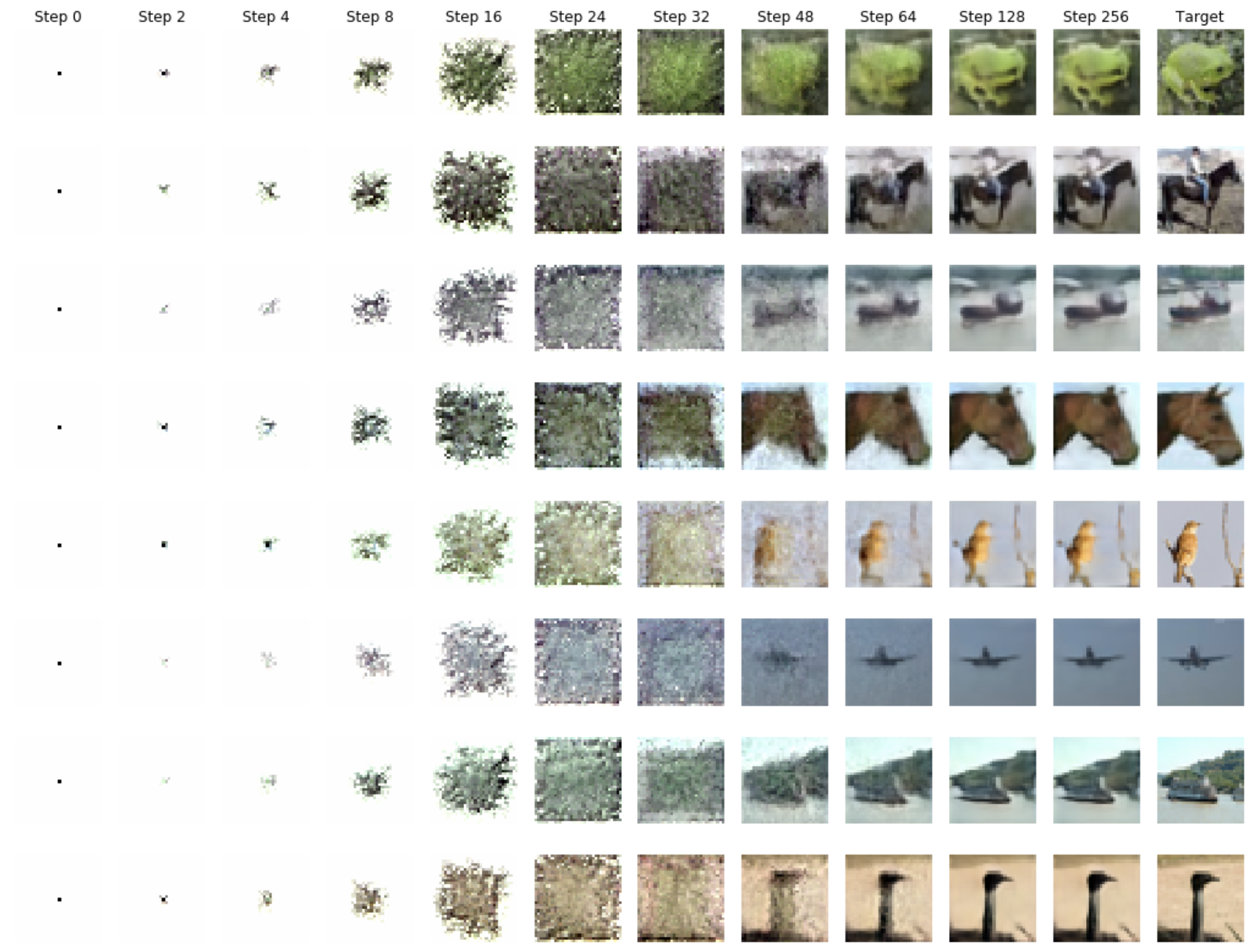}
        \caption{CIFAR}
    \end{subfigure}
    \caption{Examples of images' growing process.}\label{Fig:steps}
\end{figure}

\begin{figure}[t!]
    \centering
    \begin{subfigure}[t]{0.47\linewidth}
        \includegraphics[width=.99\linewidth]{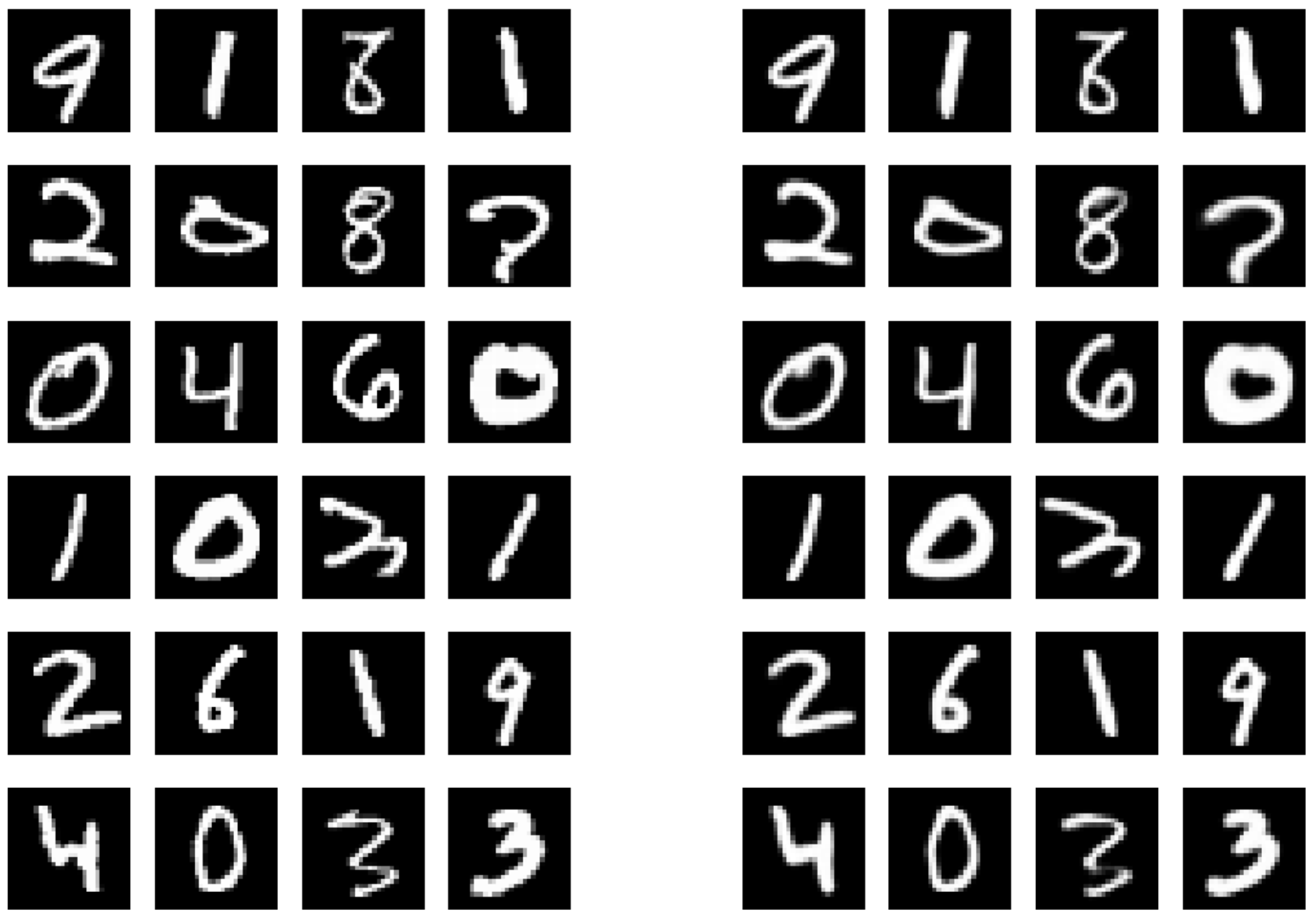}
        \caption{MNIST}
    \end{subfigure}
    \hfill
    \begin{subfigure}[t]{0.47\linewidth}
        \includegraphics[width=.99\linewidth]{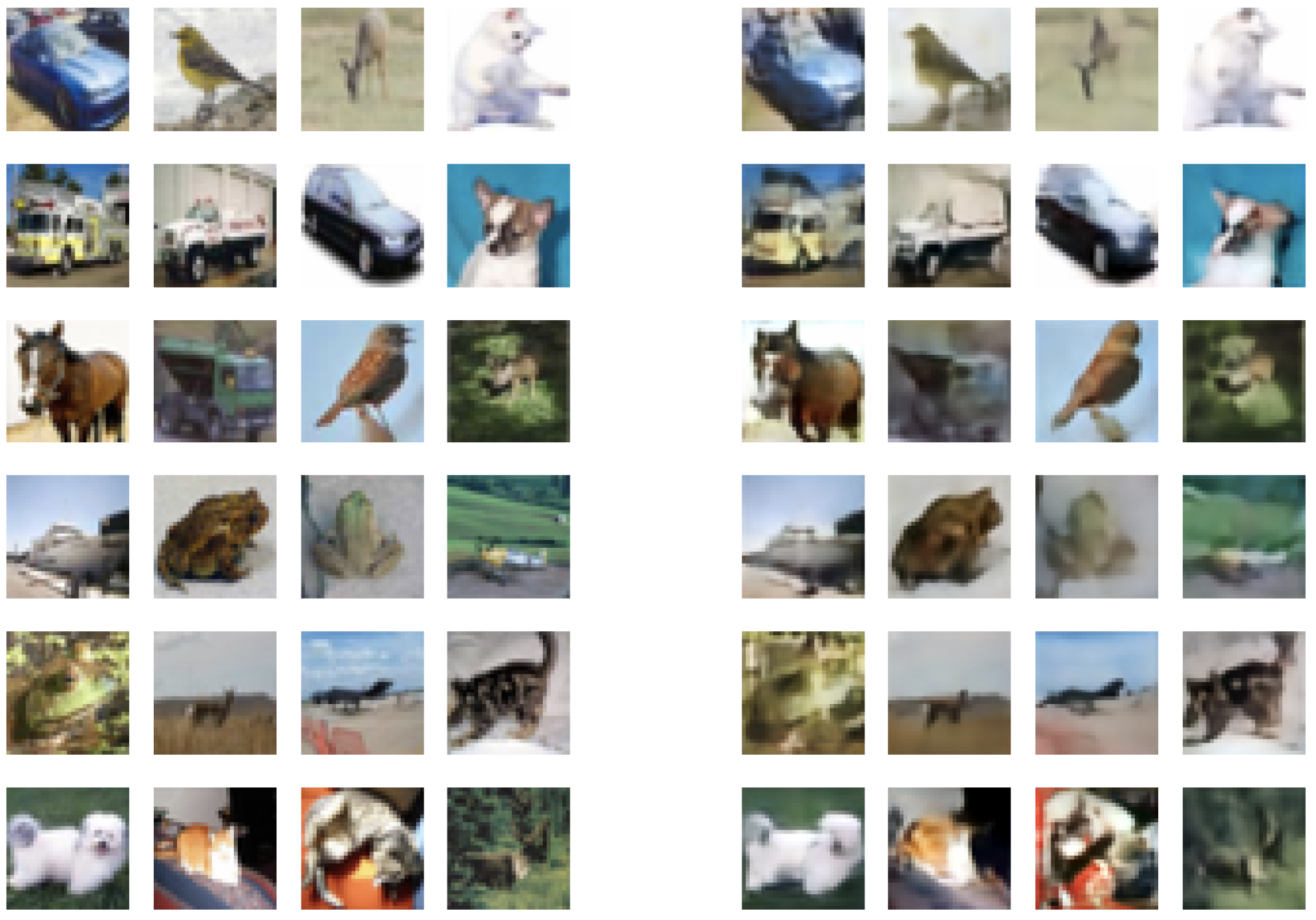}
        \caption{CIFAR}
    \end{subfigure}
    \caption{An illustration of reconstructing images in the MNIST and the CIFAR dataset (test set). The ground truth images are shown on each's left, and the model's outputs are on the right.}\label{Fig:reconstruct}
\end{figure}

\begin{figure}[t!]
    \begin{subfigure}[t]{0.445\linewidth}
        \centering
        \includegraphics[width=.95\linewidth]{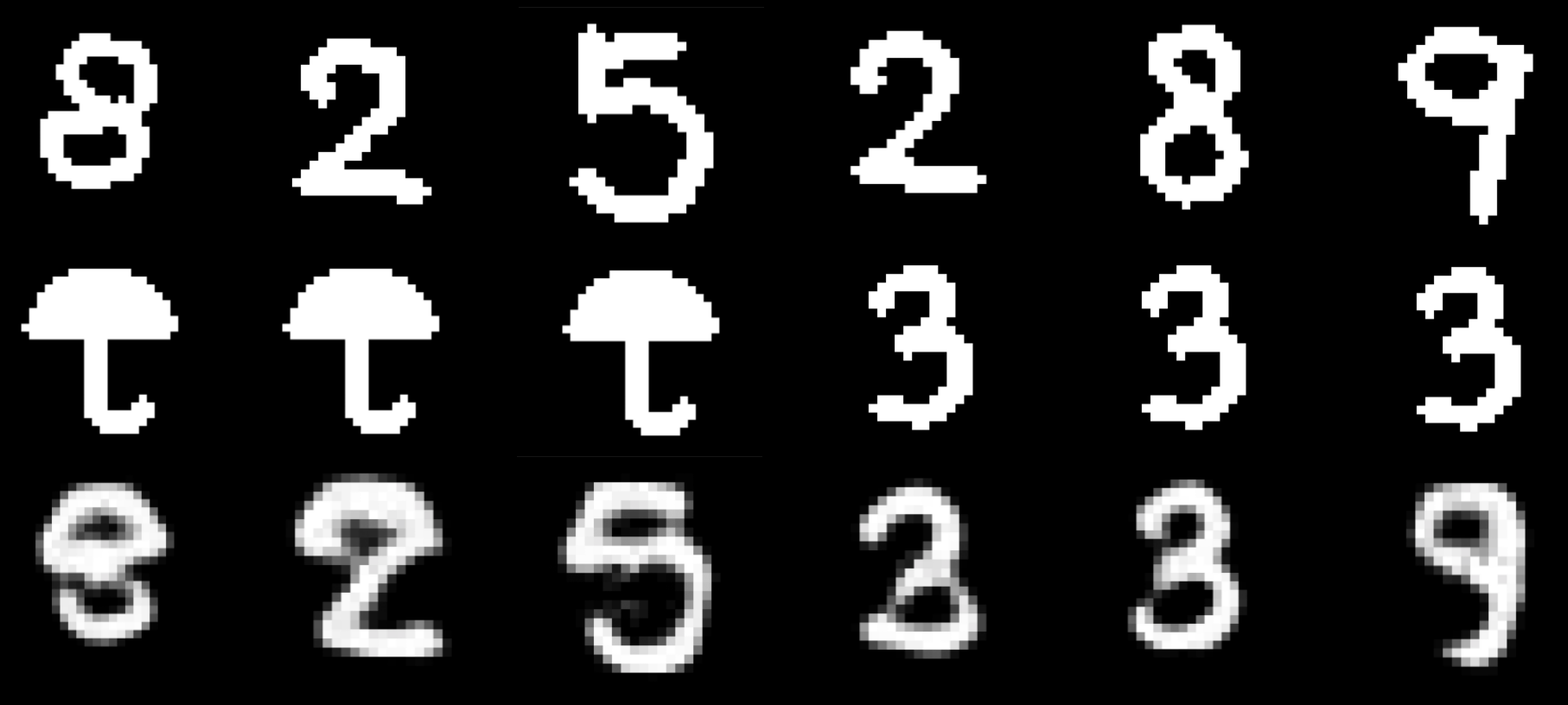}
        \caption{An illustration of style fusion using the MNIST dataset (test set). The first two rows are the sources, while the third row is the output of fusing the source images. \zhecheng{For example, in the 1st column, the output is a combination of ``8'' and the umbrella.}}\label{Fig:num_style}
    \end{subfigure}
    \hfill
    \begin{subfigure}[t]{0.545\linewidth}
        \centering
        \includegraphics[width=.95\linewidth]{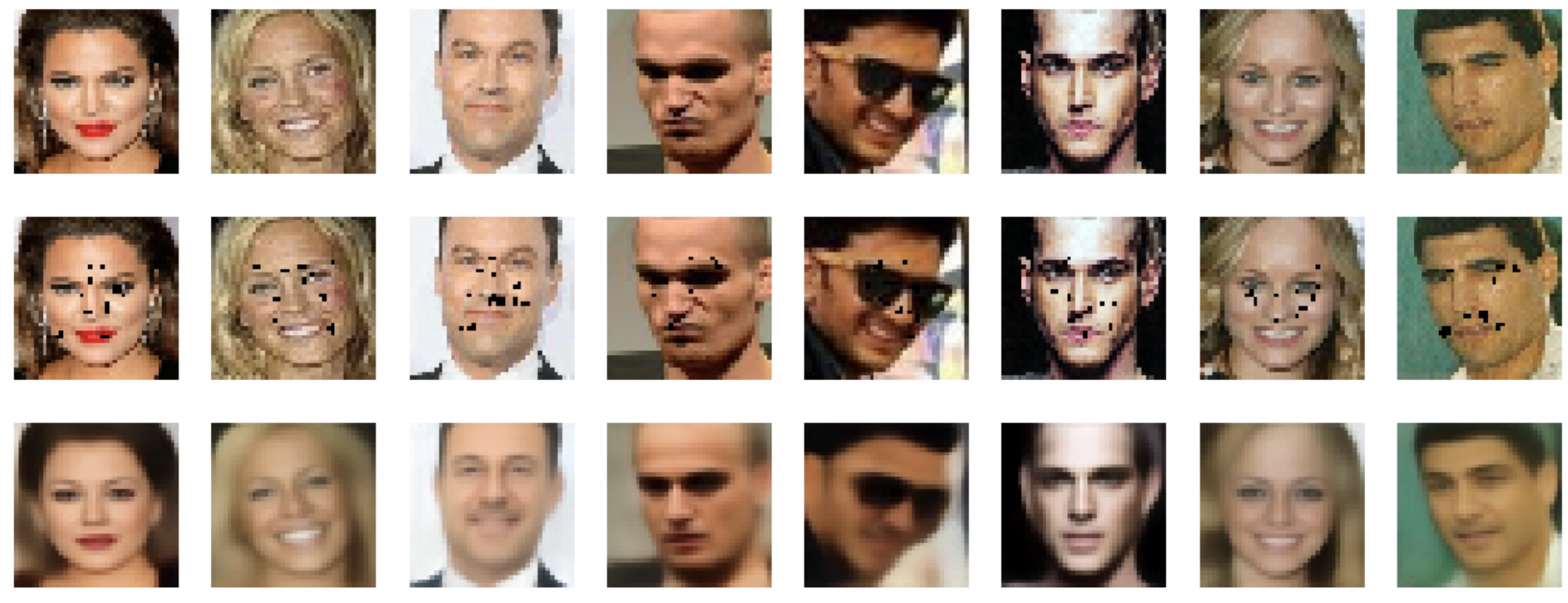}
        \caption{An illustration of human face repair using the celebA dataset (test set). The first row is the ground truth image, the second row is the result of partially damaging the original picture, and the third row is the output of the model by taking the images from the second row as input.}\label{Fig:vgg_repair}
    \end{subfigure}
    \caption{Style fusion and image repair.}
\end{figure}

Before presenting the results, we first briefly describe the image generating process. The canvas is firstly initiated \zhecheng{with} 
all black \zhecheng{pixels} except the pixel at the center \zhecheng{which} is white. The VAE-NCA model takes the raw images as inputs and the information is mapped to a hidden high-dimensional space (a vector). The update rules are then reconstructed from this vector, which is then iterated on the original canvas for several steps. The two models shown in figure \ref{Fig:reconstruct} illustrate that a single model trained with the MNIST dataset or the CIFAR dataset is able to reconstruct the raw images. Figure \ref{Fig:steps} shows how the pattern on the canvas changes as the iteration progresses. Furthermore, if we have two sets of rules based on different images, we can obtain an image \zhecheng{fusing} 
the two source images by updating the canvas using two rules in turn, where the result is shown in figure \ref{Fig:num_style}. Moreover, as long as the defaced picture is used as input during training, and the complete picture is used to calculate the loss, the algorithm can also be used to repair the defacement as shown in the figure \ref{Fig:vgg_repair}. \zhecheng{As the generation is by iteration, VAE-NCA can also be potentially used for video generation, which is part of our future work.}

\section{Ethical Implications}

Six years ago, Generative Adversarial Networks (GANs) set off a revolution in the field of deep learning. This revolution produced some major technological breakthroughs. Both academia and industry have begun to accept and welcome the arrival of GAN. The rise of GAN is inevitable. The power of image generation algorithms is that their learning process is unsupervised. This means that they do not need to label the data, which makes them powerful, as the work of data labeling is time consuming. In addition, with the continuous \zhecheng{evolution }
of technology, deep learning algorithms have been able to complete many promising applications such as generating high-quality images, image enhancement, generating images from text, converting images from one domain to another, changing the appearance of faces, and etc. Apparently, the above list is far from enough.

However, in 2017, the presence of Deepfake began to spread across Internet. People use Deepfake to disrupt political elections, discredit public figures, and directly lead to the proliferation of involuntary pornography, thereby gradually trigger a crisis of social trust. With the \zhecheng{decreasing }
implementation cost, Deepfake has penetrated \zhecheng{increasingly} deeper 
in all aspects of social life. From the perspective of the public, the current public attitude towards Deepfake is going to two extremes, one is falling into the vortex of Deepfake entertainment, and the other is standing on the opposite side of technology and rejecting all deep learning-related technologies.

The current academic research on NCA is not very in-depth. In the short term, it is difficult to cause technical abuse and social thinking similar to Deepfake. In the long run, if more scholars are willing to invest in this direction, it will only be a matter of time before using NCA to generate more refined images or complete more difficult tasks. In other words, we cannot rule out the possibility of abusing NCA. However, we also believe that technological innovation itself should be encouraged morally. The social trust crisis caused by Deepfake \zhecheng{should be temporary, } 
because as its abuse occurs frequently, more accurate detection models, better \zhecheng{legislation, }
and a more sound social moral system will be established. Justice will be late, but it will never be absent. We can’t comment on the practical significance of our work, but as the old Chinese saying goes, ``throwing bricks to attract jade'', if we want to find out how to solve similar problems in the future, we may eventually need technology to deal with technology.

\bibliographystyle{achemso}
\bibliography{nips}

\providecommand{\latin}[1]{#1}
\makeatletter
\providecommand{\doi}
  {\begingroup\let\do\@makeother\dospecials
  \catcode`\{=1 \catcode`\}=2 \doi@aux}
\providecommand{\doi@aux}[1]{\endgroup\texttt{#1}}
\makeatother
\providecommand*\mcitethebibliography{\thebibliography}
\csname @ifundefined\endcsname{endmcitethebibliography}
  {\let\endmcitethebibliography\endthebibliography}{}
\begin{mcitethebibliography}{5}
\providecommand*\natexlab[1]{#1}
\providecommand*\mciteSetBstSublistMode[1]{}
\providecommand*\mciteSetBstMaxWidthForm[2]{}
\providecommand*\mciteBstWouldAddEndPuncttrue
  {\def\EndOfBibitem{\unskip.}}
\providecommand*\mciteBstWouldAddEndPunctfalse
  {\let\EndOfBibitem\relax}
\providecommand*\mciteSetBstMidEndSepPunct[3]{}
\providecommand*\mciteSetBstSublistLabelBeginEnd[3]{}
\providecommand*\EndOfBibitem{}
\mciteSetBstSublistMode{f}
\mciteSetBstMaxWidthForm{subitem}{(\alph{mcitesubitemcount})}
\mciteSetBstSublistLabelBeginEnd
  {\mcitemaxwidthsubitemform\space}
  {\relax}
  {\relax}

\bibitem[Mordvintsev \latin{et~al.}(2020)Mordvintsev, Randazzo, Niklasson, and
  Levin]{mordvintsev2020growing}
Mordvintsev,~A.; Randazzo,~E.; Niklasson,~E.; Levin,~M. Growing Neural Cellular
  Automata. \emph{Distill} \textbf{2020},
  https://distill.pub/2020/growing-ca\relax
\mciteBstWouldAddEndPuncttrue
\mciteSetBstMidEndSepPunct{\mcitedefaultmidpunct}
{\mcitedefaultendpunct}{\mcitedefaultseppunct}\relax
\EndOfBibitem
\bibitem[McCulloch and Pitts(1943)McCulloch, and Pitts]{mcculloch1943logical}
McCulloch,~W.~S.; Pitts,~W. A logical calculus of the ideas immanent in nervous
  activity. \emph{The bulletin of mathematical biophysics} \textbf{1943},
  \emph{5}, 115--133\relax
\mciteBstWouldAddEndPuncttrue
\mciteSetBstMidEndSepPunct{\mcitedefaultmidpunct}
{\mcitedefaultendpunct}{\mcitedefaultseppunct}\relax
\EndOfBibitem
\bibitem[Li and Yeh(2002)Li, and Yeh]{li2002neural}
Li,~X.; Yeh,~A. G.-O. Neural-network-based cellular automata for simulating
  multiple land use changes using GIS. \emph{International Journal of
  Geographical Information Science} \textbf{2002}, \emph{16}, 323--343\relax
\mciteBstWouldAddEndPuncttrue
\mciteSetBstMidEndSepPunct{\mcitedefaultmidpunct}
{\mcitedefaultendpunct}{\mcitedefaultseppunct}\relax
\EndOfBibitem
\bibitem[Wu and Kareem(2011)Wu, and Kareem]{wu2011modeling}
Wu,~T.; Kareem,~A. Modeling hysteretic nonlinear behavior of bridge
  aerodynamics via cellular automata nested neural network. \emph{Journal of
  Wind Engineering and Industrial Aerodynamics} \textbf{2011}, \emph{99},
  378--388\relax
\mciteBstWouldAddEndPuncttrue
\mciteSetBstMidEndSepPunct{\mcitedefaultmidpunct}
{\mcitedefaultendpunct}{\mcitedefaultseppunct}\relax
\EndOfBibitem
\bibitem[Lauret \latin{et~al.}(2016)Lauret, Heymes, Aprin, and
  Johannet]{lauret2016atmospheric}
Lauret,~P.; Heymes,~F.; Aprin,~L.; Johannet,~A. Atmospheric dispersion modeling
  using Artificial Neural Network based cellular automata. \emph{Environmental
  Modelling \& Software} \textbf{2016}, \emph{85}, 56--69\relax
\mciteBstWouldAddEndPuncttrue
\mciteSetBstMidEndSepPunct{\mcitedefaultmidpunct}
{\mcitedefaultendpunct}{\mcitedefaultseppunct}\relax
\EndOfBibitem
\end{mcitethebibliography}

\newpage

\begin{appendices}
\section{Method Detail}

The development of modern artificial intelligence is inextricably linked to bionics and biology. In the early 1940s, researchers have built the most primitive artificial neural network [\cite{mcculloch1943logical}] based on the neuron structures in living bodies. Afterwards, as the computing power continued to increase, the experimental methods also changed dramatically. More recently, with the continuous developement of convolutional networks, recurrent neural networks, and various training methods, artificial intelligence has entered \zhecheng{into} a new stage of rapid evolution.

On the other hand, because the practical value has not yet been fully explored, the application of another kind of artificial intelligence model, cellular automata, both in academia and industry, has not been \zhecheng{used} as extensively as deep learning. Cellular automata (CA) is sometimes described as game of life, which is a series of widely-used modeling methods. It refers to a dynamic system defined by the following four key components:

\begin{itemize}

    \item Discrete states: The most basic component of CA are cells. Each cell has its own independent discrete and finite state and is usually expressed in the form of numbers or vectors. Taking Conway’s game of life as an example, the state set is $\{0,1\}$.
    
    \item Discrete space: Cells are distributed in a discrete Euclidean space, where the number of dimensions and the layout of the cells could vary.
    Yet in most cases, cells are assumed to be qualitatively identical.
    
    \item Discrete dynamics: The dynamics of the system controlled by the ``rules'' is usually noncontinuous. The starting pattern (cell states) of the system is called ``configuration'', and the states of the cells at the each time step depend on the states at previous steps.
    
    \item Local interactions: The way the system evolves depends on the rules of interaction between cells. Other cells within the local space which may be interacted with are defined as the ``neighborhood''. While the evolution for each cell only takes place based on local information, a large number of cells make the evolution of the entire dynamic system happen through interactions, and hence form a dynamic effect globally.

\end{itemize}

While the mechanism may be simple to understand, designing new CAs is not easy. As a result, some researchers began to explore solutions that combine deep learning methods with cellular automata. These methods are applied to land-use simulation [\cite{li2002neural}], aerodynamics [\cite{wu2011modeling}], atmospheric dispersion [\cite{lauret2016atmospheric}], and etcetera. As for the way of interpreting the meaning of the parameters within neural cellular automata (NCA) models, there is no need to stick to the concept of "neural network". A more recent study [\cite{mordvintsev2020growing}] on NCA showed that these parameters to the models can also be likened as genes to creatures. The study discussed a series of methods for designing and training NCAs, and achieved good results in the tasks of growing, persisting, and regenerating a single emoji image. Similar to real cell collectives, the trained model is very robust to perturbations. However, even if it is not difficult to obtain a CA with excellent performance within an acceptable training time using the algorithm described in the study, it takes at least a few minutes to finish such a process. If a certain method can be used to accelerate the process and output a similar NCA model, its application range will be greatly expanded.

\begin{figure}[t!]
    \centering
    \includegraphics[width=.99\linewidth]{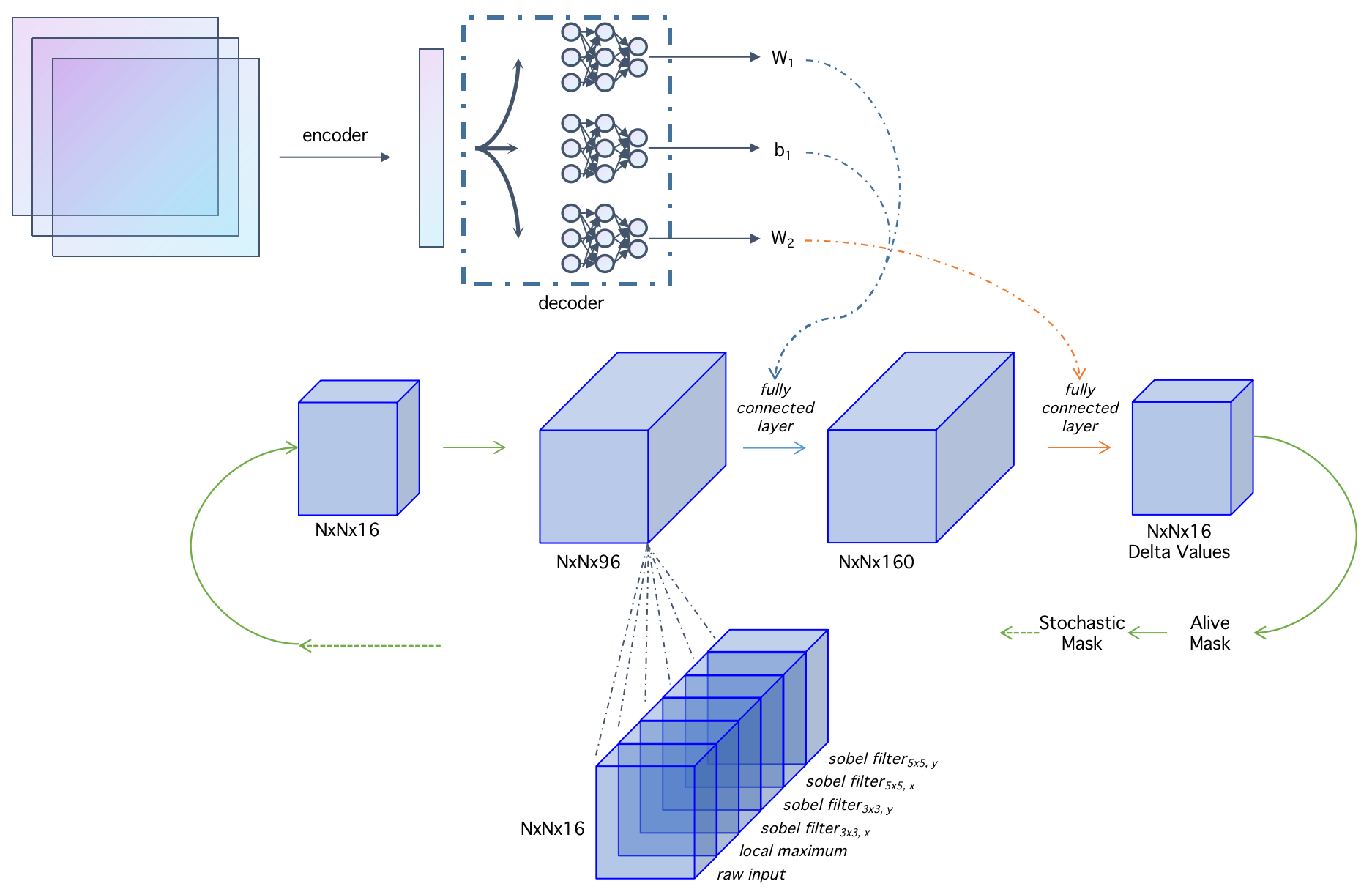}
    \caption{An illustration of model's structure.}\label{Fig:structure}
\end{figure}

Our model combines the characteristics of VAE and NCA. Very similar to VAE, the function of the encoder part is to map the picture to a high-dimensional space using a set of convolutional layers, but it does not directly output the specific values of each dimension, but in the form of mean and standard deviation. By establishing normal distributions based on these, and randomly extracting values from them, the encoder would output a set of implicit values representing image information. The vector composed of these values will be the input of the decoder, which has a dimension of $E$. The decoder is a simple structure composed of a series of multilayer perceptrons (MLPs). Each MLP takes the output of the encoder as input, and the output values will be then used as the neural network parameters of the NCA.

Unlike the CA described above, in the NCA model demonstrated in this article, the state of each cell is represented by a vector. Therefore, the state set does not only contain zeros and ones, but has more possibilities. The first four dimensions of each vector represent the RGBA values, while the remaining dimensions are used to represent some hidden information. Therefore, a huge tensor composed of a large number of cells can not only be used to visually display the complex patterns of an image, but also can be used to store the temporal information retained during the iteration process.

Before the iteration starts, the values in all cells in the canvas are initialized to zero, except for the cell in the center of the canvas, which the first four values in its vector are initialized as one. As shown in figure \ref{Fig:structure}, the model firstly uses different sizes of sobel filters and a local maximum filter to process the original values, where the window size of the local maximum is $w$. Then through the mapping of two fully connected layers (using ReLU as the activation function, where the dimension of the hidden layer is $d$), it will finally get a set of updated values with the same shape as the input tensor. The parameters used are the output values of the decoder. The addition of the updated values and the original values are the results. The values on the canvas will be continuously be updated by repeating this process. We train the model so that after $T$ iterations, the states' first four dimensions are close to the encoder's input image as close as possible. The mean square loss is used to measure the loss, and the Adam optimizer is used to optimize the parameters.

In the MNIST dataset examples, each cell is composed of a 8-dimension vector. $w=3$, $E=32$, $d=64$, and $T=64$. $3\times 3$ Sobel filters and $7\times 7$ Sobel filters are used to extract features.

In the CIFAR dataset examples, each cell is composed of a 16-dimension vector. $w=3$, $E=1024$, $d=160$, and $T=128$. $3\times 3$ Sobel filters and $5\times 5$ Sobel filters are used to extract features.

In the human face repair examples, each cell is composed of a 24-dimension vector. $w=3$, $E=1024$, $d=256$, and $T=160$. $3\times 3$ Sobel filters, $5\times 5$ Sobel filters, and $9\times 9$ Sobel filters are used to extract features.

\end{appendices}

\end{document}